\documentclass{article}
\ifdefined\pdfminorversion\pdfminorversion=7\fi
\usepackage{amsmath} 
\usepackage{amssymb}
\usepackage{booktabs}   
\usepackage{graphicx}   
\usepackage{amsmath}
\usepackage{pifont}
\newcommand{\cmark}{\ding{51}}
\newcommand{\xmark}{\ding{55}}
\DeclareMathOperator*{\argmin}{arg\,min}
\usepackage{xcolor}
\definecolor{DeepFigBlue}{RGB}{0,55,160}
\definecolor{PRISMGreen}{HTML}{298E76}
\definecolor{pink}{HTML}{FD7979}

\usepackage[final]{corl_2026} 
\usepackage{xcolor}
\definecolor{cfyellow}{RGB}{255,245,170}
\definecolor{tile}{HTML}{0097A7}
\newcommand{\tok}[1]{\texttt{\char60#1\char62}}

\usepackage{graphicx}
\usepackage{subcaption}
\usepackage{booktabs}
\usepackage{array}
\title{Counterfactual Video Generation Enables\\Scalable Humanoid Loco-Manipulation}

\newcommand{\colead}{\textsuperscript{\(\dagger\)}}

\author{
\begin{tabular}{@{}c@{}}
Zihan Wang\textsuperscript{1,2}
\quad
Zhen Wu\textsuperscript{1}
\quad
Pieter Abbeel\textsuperscript{1,2}\colead
\quad
Rocky Duan\textsuperscript{1}\colead
\quad
Jitendra Malik\textsuperscript{1,2}\colead
\\[0.4em]
Carmelo Sferrazza\textsuperscript{1}\colead
\quad
C. Karen Liu\textsuperscript{1,4}\colead
\quad
Guanya Shi\textsuperscript{1,3}\colead
\quad
Angjoo Kanazawa\textsuperscript{1,2}\colead
\\[0.55em]
{\normalfont\mdseries\small
\textsuperscript{1}Amazon FAR
\quad
\textsuperscript{2}UC Berkeley
\quad
\textsuperscript{3}Carnegie Mellon University
\quad
\textsuperscript{4}Stanford
\quad
\(\dagger\) FAR Team Co-Leads
}
\end{tabular}
}
\begin{document}
\maketitle

\begin{center}
\begin{figure}[ht]
  \centering
  \vspace{-0.99cm}
  \includegraphics[width=0.93\linewidth]{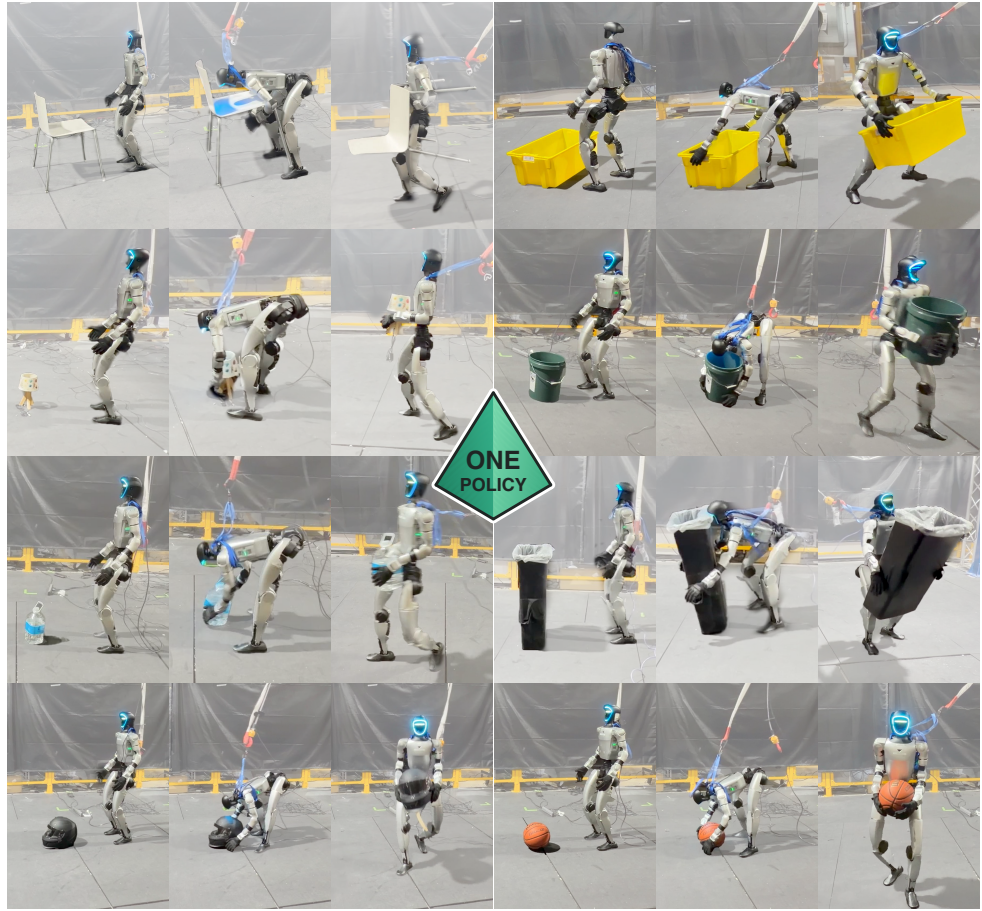}

\caption{
\textit{PRISM} leverages video-to-video (V2V) generation to expand a few real videos into diverse counterfactual interactions---interactions that did not occur in the source videos but could have occurred with different objects. Reconstruction and retargeting yield physically plausible robot--object trajectories for training a unified depth-based humanoid policy that picks up, carries, and drops diverse objects zero-shot in real world deployment. Project website: \href{https://prism-real2sim2real.github.io/}{\textcolor{PRISMGreen}{prism-real2sim2real.github.io}}.
}
\vspace{-0.7cm}
\label{fig:pre_teaser}
\end{figure}
\end{center}%

\begin{abstract}
Teaching humanoids loco-manipulation skills, such as carrying diverse objects, via visual imitation is a promising path toward generalist robots. However, collecting diverse, high-quality interaction videos, such as clips that clearly show a person’s full body and unoccluded interactions with objects, poses a practical barrier to scaling this approach. 
We propose PRISM, a real-to-sim-to-real framework that overcomes this limitation by amplifying a handful of real videos into a large, diverse training set. PRISM first generates hundreds of diverse ``counterfactual'' human–object interaction videos via video-to-video (V2V) generation from a few exemplar real videos. Our contact-anchored real-to-sim pipeline then reconstructs both human and object motions, retargeting this imperfect video data into physically plausible trajectories. The intra-class variability across these counterfactual videos lets us train a single policy that generalizes to unseen objects within each category. We demonstrate the full pipeline by deploying this policy on a real robot without any real-world fine-tuning. Using only onboard depth observations, our humanoid picks up, carries, and drops objects—including boxes, barrels, bins, and balls---across novel instances, sizes, and initial configurations. 
    
\end{abstract}

\vspace{-0.3cm}
\keywords{Loco-manipulation, Real-to-Sim-to-Real, Humanoid} 

\section{Introduction}

How can humanoids learn to interact with the diverse objects encountered in everyday life? Visual imitation learning offers a promising route: human demonstrations provide examples of the coordinated whole-body motions needed to approach, lift, carry, and drop various objects. Recent real-to-sim-to-real pipelines~\citep{videomimic} have enabled humanoids to acquire contextual whole-body locomotion skills directly from human videos. These advances suggest a path toward generalist humanoids that learn a broad repertoire of locomotion and manipulation skills from internet-scale video data.

However, acquiring diverse, high-quality human--object interaction videos remains a practical bottleneck to scaling visual imitation. Internet videos are abundant, but their content and framing are shaped by human viewing preferences. Filtering Internet videos to find demonstrations that clearly show the person's full body and how they interact with objects is therefore costly and impractical at scale. Yet, such interaction data \textit{does} exist implicitly in modern video generative models~\citep{ye2026world}, whose learned priors over human motion and interactions can be used to synthesize diverse training videos.


To address this data bottleneck, we propose \textit{PRISM}, a real-to-sim-to-real framework that uses video-to-video (V2V) generation to expand a few real videos into diverse human--object interactions. We call these generated clips ``counterfactual videos'': they depict interactions that did not occur in the source videos but could have occurred with different objects. From only four real-world videos of humans carrying boxes, we generate 256 counterfactual videos depicting the same task with boxes, balls, bins, and barrels of varying geometries and initial configurations. Our contact-anchored real-to-sim pipeline reconstructs human motion, the static scene, and dynamic object geometry and motion in a unified world frame, then retargets these imperfect reconstructions into physically plausible humanoid--object trajectories. With these trajectories, we train a single depth-based policy that picks up, carries, and drops unseen instances of these categories zero-shot in the real world, validating the full pipeline. Our policy also generalizes to categories absent from the generated videos (Fig.~\ref{fig:pre_teaser}).

The core technical challenge is to obtain physically plausible robot demonstrations despite errors introduced by both counterfactual video generation and monocular reconstruction. Our key insight is to use contact signal as a shared constraint across reconstruction, retargeting, and policy learning. In the first reconstruction stage, human--object contact couples the two motions: human motion guides the object trajectory, while object contact helps correct errors in the reconstructed human pose. During the retargeting stage, sparse contact anchors specify where robot end-effectors should contact the object, guiding the refinement of noisy reconstructions into physically plausible robot--object trajectories while accommodating morphological differences. During policy learning, these anchors define contact rewards for a privileged teacher policy that co-tracks robot and object motion. We then distill it into a depth-based policy with joystick control for zero-shot sim-to-real deployment.

We summarize our contributions as follows: (1) We introduce a new learning paradigm that uses counterfactual video generation for scaling human--object interaction experience. (2) We propose a contact-anchored real-to-sim and retargeting pipeline that turns monocular reconstructions into physically plausible robot--object demonstrations. (3) We demonstrate the first real-to-sim-to-real pipeline that trains a unified whole-body humanoid policy that generalizes across diverse objects and enables joystick-controlled pick-up, carry, and drop behaviors from onboard depth observations.



\section{Related Work}
\vspace{-0.15cm}
\label{sec:related}

\paragraph{Humanoid Loco-Manipulation.}




Humanoid loco-manipulation has been widely studied in both computer graphics~\citep{wu2024human, li2023object} and robotics~\citep{weng2025hdmi, omniretarget,wang2026vlk}, where the goal is to coordinate whole-body locomotion and object interaction simultaneously. Some humanoid systems~\citep{weng2025hdmi} learn interaction skills by tracking human motion references, but require dense reference motions or object trajectories during inference, limiting generalization to unseen objects and interaction configurations. More recent methods~\citep{lin2026lessmimic, he2026ultra, wang2026generalizing} reduce this dependency by learning more autonomous interaction policies from sparse goals or interaction-centric representations. Similarly, PRISM trains a unified whole-body humanoid policy that enables zero-shot joystick-controlled pick-up, carry, and drop behaviors from onboard depth observations, without reference or motion-capture systems during deployment.

\vspace{-0.1cm}
\paragraph{Learning from Human Video.}



Recent works learn humanoid skills from human videos~\cite{weng2025hdmi, mao2024learning}, including terrain-aware locomotion via joint human--scene reconstruction~\cite{videomimic} and dynamic object interactions~\cite{weng2025hdmi, wang2026humanx}. However, capturing or curating diverse, high-quality interaction videos at scale remains challenging. PRISM addresses this complementary challenge by expanding a few real videos into diverse human--object interactions through video-to-video generation. With these data, we train a single policy that picks up, carries, and drops diverse objects zero-shot in the real world.

\vspace{-0.1cm}
\paragraph{Video Model as Data++.} 


Recent works have explored video generation for robot learning under two main paradigms. 
One line uses video models as inference-time planners or action generators, synthesizing future visual rollouts to guide closed-loop manipulation~\cite{chen2025large, bharadhwaj2024gen2act, ye2026world}.
While such methods provide flexible visual reasoning, they require expensive test-time generation and may suffer from an executability gap between plausible videos and physically feasible robot actions. 
Another line uses video models offline to synthesize robot demonstrations or augment robot datasets before policy learning~\cite{patel2025robotic, jang2025dreamgen}. 
PRISM follows the offline-generation paradigm, but differs in generating counterfactual human interaction videos through grounded V2V generation rather than directly generating robot videos from text or images. 
By conditioning on real demonstrations, V2V preserves realistic motion, lighting, and affordance-aware contacts, while a simple unified prompt can scale one exemplar into diverse object categories, poses, and intra-class variations by multiple sampling.

\vspace{-0.15cm}
\section{Real-to-Sim Data Acquisition}
\vspace{-0.15cm}

Given a monocular video depicting a human interacting with a static scene and a dynamic object, our goal is to recover camera parameters $(K, \{T_i^c\})$, a static scene mesh $M_s$, a dynamic object with its metric mesh $M_o$ and its per-frame poses $\{T_i^o\}$, and 4D SMPL-X~\citep{SMPL-X:2019} motion, all in a unified world frame. Then we retarget the recovered data to humanoid--object trajectories for policy learning.

\vspace{-0.1cm}
\subsection{Counterfactual Interaction Video Generation}
\vspace{-0.1cm}
Scaling real-to-sim learning with online videos remains challenging. Although internet videos are abundant, their content and framing are shaped by human viewing preferences. Filtering for diverse clips with clear full-body views, visible human–object interactions, and stable viewpoints is therefore costly and impractical at scale. We use video-to-video (V2V) generation to expand a small set of suitable real videos into diverse \emph{counterfactual interaction videos}---interactions that did not occur in the source videos but could have occurred with different objects. Given a seed video and a category-level text prompt, we ask the model to replace the manipulated object while preserving the original background, lighting, camera viewpoint, and coarse task structure. This video conditioning grounds generation in a real scene and task while allowing object and human behavior to vary.

Importantly, the counterfactual variation is not limited to object geometry or appearance. When the object category, size, pose, or placement changes, the object's manipulation affordances also change, and the human behavior adapts accordingly: the generated person may bend lower, adjust hand spacing, adapt the contact strategy based on the object's affordances, or carry the object differently. Thus each generated clip provides an object-conditioned interaction strategy, rather than merely pairing the same human motion with a new object. This allows PRISM to obtain behavior-level variations in interaction data that are difficult to hard-code through geometry-level augmentation alone. Learning such adaptations from scratch would require exhaustive task-specific RL reward design; PRISM instead uses the video model as an offline prior over plausible human adaptations.

\begin{figure}[t]
\begin{center}
\includegraphics[width=0.9\linewidth]{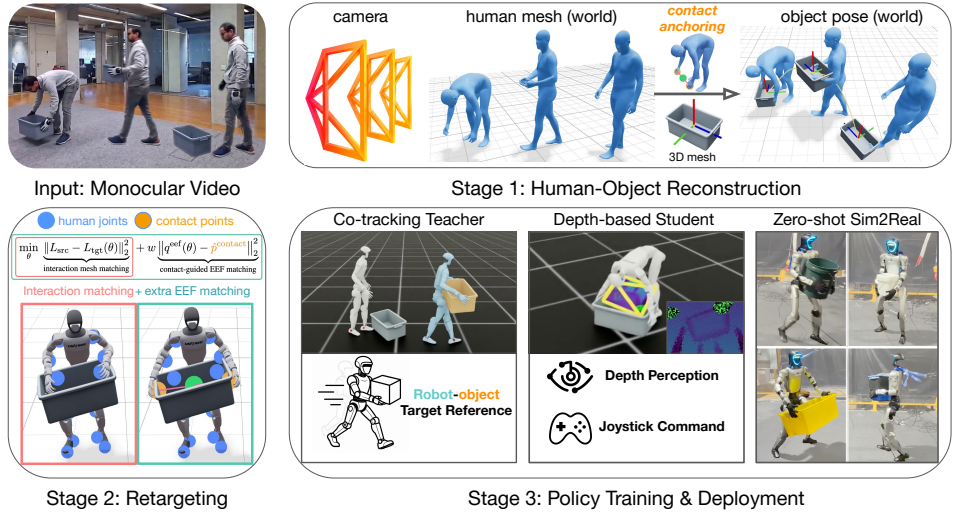}
\end{center}
\caption{\textbf{PRISM Real-to-sim Overview.}
PRISM turns counterfactual human--object videos into deployable humanoid loco-manipulation skills.
It reconstructs the camera, human motion, object geometry, and 6D object motion in a shared world frame, using  \textcolor{orange}{contact points} to constrain object pose optimization under monocular ambiguity.
The same anchors are used in retargeting to preserve interaction phases and match robot end-effectors to intended \textcolor{orange}{contact points}.
The retargeted demonstrations train a privileged co-tracking teacher, which is distilled into a depth-based student policy conditioned on onboard depth and joystick commands for zero-shot sim-to-real deployment. 
}
\vspace{-0.5cm}
\label{fig:method}
\end{figure}


In practice, we record four real-world seed videos and use each as a grounded template for V2V generation. We prompt SeedDance 2.0~\citep{seedance2026seedance} to replace the manipulated object with a box, bin, barrel, or ball while preserving the original background, lighting, camera viewpoint, and temporal continuity (Appendix~\ref{sec: cf-gen}). Category-level prompts allow the model to vary object geometry, appearance, and pose while adapting human behavior accordingly. With 16 samples per category, each seed yields 64 counterfactual videos, expanding four real recordings into 256 videos for our real-to-sim pipeline.

\vspace{-0.15cm}
\subsection{Contact-Anchored Real-to-Sim}
\vspace{-0.1cm}
\label{contact-3.2}
To imitate human motion from monocular video, we must disentangle it from camera-induced image motion and recover it in a consistent world coordinate frame. CRISP~\citep{wang2025crisp} integrates an human mesh recovery (HMR) network with visual SLAM to reconstruct a metrically-consistent human--scene--camera representation from monocular video, including camera intrinsics $K \in \mathbb{R}^{3 \times 3}$, per-frame camera poses $T_i = [R_i \mid t_i] \in SE(3)$, a metric-scale scene point cloud $\tilde{P}$, and temporally aligned 4D human motion in a unified world coordinate frame. We use it as backend and extend it to reconstruct both the geometry and motion of dynamic objects. While CRISP~\citep{wang2026contact} focuses on human motion and the surrounding environment, imitating human--object interactions additionally requires disentangling dynamic object motion from camera motion. We use it as our backend and extend it to recover both the geometry and motion of dynamic objects in the same world coordinate frame.


\vspace{-0.2cm}
\paragraph{Object Geometry and Motion Reconstruction.} Given dynamic object masks from SAM 2~\citep{ravi2024sam2}, we reconstruct the object geometry with SAM3D~\citep{sam3dteam2025sam3d3dfyimages}, together with the camera poses and metric depth from CRISP, yielding an object mesh $\mathcal{M}_o$ and initial world pose $T_o^{t=0}\in SE(3)$. Rather than tracking the object independently with a visual 6D tracker such as FoundationPose~\citep{wen2024foundationposeunified6dpose}, which is sensitive to hand and body occlusions in monocular video (Table~\ref{tab:ablation}), we exploit human--object contact to regularize both motions: human motion provides a strong prior for the object trajectory, while the object in turn constrains inaccurate human joint estimates. For pick--carry--drop interactions, the object remains supported by the scene outside contact and follows the human during stable contact.

We detect contact from human motion clues, avoiding manual contact annotations~\citep{weng2025hdmi, wang2026humanx}. The detected contact points also regularize human motion during IK by replacing inaccurate joint targets with contact-consistent constraints. During each contact phase $[t_1,t_2]$, we anchor the object to the SMPL-X palm using the palm-relative transform at $t_1$, held fixed throughout the phase:
\[
T_o^t
=
T_{\mathrm{palm}}^t T_{\mathrm{palm}\rightarrow o}
=
T_{\mathrm{palm}}^t
\left(T_{\mathrm{palm}}^{t_1}\right)^{-1}
T_o^{t_1},
\qquad
T_{\mathrm{palm}\rightarrow o}
=
\left(T_{\mathrm{palm}}^{t_1}\right)^{-1}
T_o^{t_1},
\qquad
t\in[t_1,t_2].
\]
Thus, human motion propagates the object trajectory during contact, while object contact, in turn, provides geometric constraints that correct errors in the reconstructed human motion.

\vspace{-0.1cm}
\paragraph{Contact-Anchored Retargeting.}
\label{contact-a}
Prior retargeting methods~\citep{omniretarget} preserve human--object relations but assume clean, physically plausible inputs. Monocular reconstructions often violate this assumption (\textcolor{pink}{pink} in Fig.~\ref{fig:method}), causing retargeting to preserve the reconstruction errors. We therefore propose contact-anchored
retargeting, which treats the reconstructed full-body human motion as a kinematic reference while
using object-frame contact anchors as the reliable interaction interface. These anchors specify where
the robot end-effectors should act on the object, allowing the retargeting solver to correct noisy
reconstructions into physically plausible robot–object motions.


For each contact phase, we derive contact anchors from the reconstructed human--object geometry in Stage 1. Specifically, we intersect the line segment  connecting the SMPL-X left and right palm centers with the object mesh. 
Following interaction-preserving retargeting~\citep{omniretarget}, we keep the constrained IK and add a contact-anchor term to the original objective. At each frame, we solve:
\begin{equation}
\label{eq:contact_anchor_retargeting}
dq_a^\star =
\argmin_{dq_a \in \mathcal{C}(q_a)}
E_{\mathrm{base}}(q_a,dq_a)
+
w_c
\left\|
x_{\mathrm{eef}}(q_a)
+
J_{\mathrm{eef}}(q_a)dq_a
-
c
\right\|_2^2 .
\end{equation}
Here $E_{\mathrm{base}}$ denotes the interaction mesh matching from~\citep{omniretarget}, $c$ is the \textcolor{orange}{target contact point} estimated from the human--object reconstruction in Fig.~\ref{fig:method}, and $x_{\mathrm{eef}}(q_a)+J_{\mathrm{eef}}(q_a)dq_a$ is the linearized end-effector position. We update $q_a \leftarrow q_a+dq_a^\star$ each frame and use it to warm-start the next frame.

One might attribute these artifacts to the reconstruction pipeline. We take a different view: imperfect reconstruction is unavoidable in monocular real-to-sim stage, as no existing pipeline can recover physically plausible human--object motions. Our contact-anchored real-to-sim system closes this gap by optimizing sufficiently close reconstructions into physically plausible robot demonstrations.

\begin{figure}[t]
\begin{center}
\includegraphics[width=0.99\linewidth]{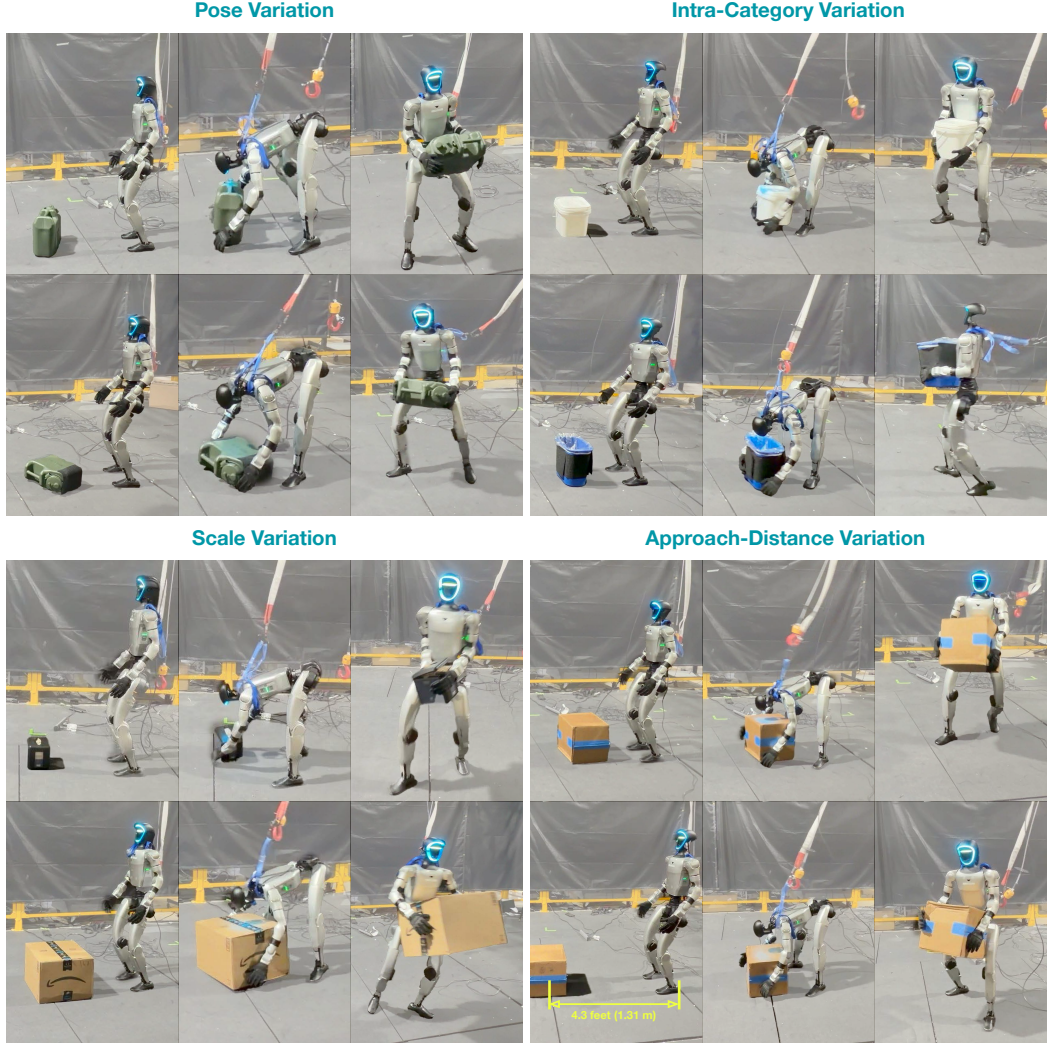}
\end{center}
\caption{\textbf{Policy rollout under real-world object variations.}
Our single unified policy zero-shot transfers to the real robot across object-pose, intra-category, scale, and approach-distance variations.
The policy is agnostic to object placement and initial pose ({\color{tile}\textit{top-left}}), handles objects with different topology and appearance within the same category ({\color{tile}\textit{top-right}}), and generalizes to large scale variations ({\color{tile}\textit{bottom-left}}).
It further adapts to object distance using onboard depth, either approaching (up to 4.3 feet!) before grasping or directly grasping when the object is within reach ({\color{tile}\textit{bottom-right}}).}
\label{fig:main_comparison}
\vspace{-0.2cm}
\end{figure}

\vspace{-0.28cm}
\section{Learning a Unified Visuomotor Interaction Policy}
\vspace{-0.15cm}
\label{sec:method}
Using the reconstructed robot-object trajectories (details are in Appendix~\ref{sec:data-details}), our goal is to train a unified humanoid policy capable of pick-up, carry, and drop behaviors across diverse objects. The policy receives onboard depth observations and joystick commands, and autonomously performs object interaction behaviors conditioned on the perceived object geometry and placement.
Similar to prior visuomotor systems~\citep{wu2026perceptive, kuang2026dex4d}, we adopt a two-stage training framework. We first train a privileged co-tracking teacher policy in simulation using full-state observations. We then distill this teacher into a unified depth-based student policy using a combination of DAgger~\citep{ross2011reduction} and reinforcement learning, enabling zero-shot sim-to-real deployment. An overview is shown in Fig.~\ref{fig:method}.

\vspace{-0.1cm}
\subsection{Training a Co-Tracking Teacher Policy}
\vspace{-0.1cm}

We formulate humanoid object interaction as a co-tracking problem, where the policy simultaneously tracks both the humanoid motion and the dynamic object trajectory reconstructed from monocular videos. 
We refer readers to our codebase for our motion-tracking training details.

\vspace{1.2cm}
\textbf{Observations.} Teacher observations include reference motion, robot proprioception, the previous action, and both current and target object states, represented by object pose and 3D bounding-box dimensions, making the policy explicitly aware of both the desired robot motion and the object state.

\textbf{Rewards and Terminations.} The reward primarily consists of robot pose tracking, object pose tracking, action rate, joint limits, and collision penalties. We additionally introduce a contact-aware interaction reward using the contact anchors from Sec~\ref{contact-a}. Specifically, we encourage the robot end-effectors to reach the desired contact positions and exceed a predefined contact-force threshold:
\begin{equation}
R_{\text{contact}, i}
=
\exp\left(-\frac{\|\mathbf{p}_{\text{eef}, i} - \mathbf{p}_{\text{target}, i}\|_2}{\sigma_{\text{pos}}}\right)
\cdot
\min\left(
\exp\left(\frac{\|\mathbf{F}_{\text{contact}, i}\|_2 - F_{\mathrm{thres}}}{\sigma_{\text{frc}}}\right),
1
\right).
\end{equation}
where $\mathbf{p}_{\text{eef}, i}$ denotes the end-effector position, $\mathbf{p}_{\text{target}, i}$ is the reconstructed target contact point, and $\mathbf{F}_{\text{contact}, i}$ is the contact force vector. This contact-aware reward stabilizes grasping and carry behaviors under noisy monocular reconstructions and substantially improves interaction quality during policy learning. We also adopt early termination and domain randomization following~\citep{omniretarget}.

\subsection{Distilling a Unified Depth-Based Student Policy}

The privileged teacher learns stable object interactions in simulation but depends on information unavailable on real hardware. We therefore distill it into a deployable depth-based student policy using only onboard perception and joystick commands, without MOCAP system or reference motion.

\textbf{Distillation.} We distill the teacher policy using a combination of DAgger and PPO objectives:
\begin{equation}
\mathcal{L} = \lambda \mathcal{L}_{\text{PPO}} + (1 - \lambda) \mathcal{L}_D,
\end{equation}
where $\mathcal{L}_D$ denotes the DAgger imitation loss and $\mathcal{L}_{\text{PPO}}$ is the RL objective. Following~\citep{wu2026perceptive}, we use a curriculum that gradually increases $\lambda$ during training and keeps $\lambda=0.9$ for the final 20K iterations.

\textbf{Observations.} The student receives proprioception, joystick commands, and onboard depth. At deployment, we estimate depth offboard from stereo images using Fast-FoundationStereo~\citep{wen2026fastfoundationstereo}. We find that this substantially reduces the sim-to-real gap in depth observations. Proprioception includes angular velocity, joint states, and the previous action. Joystick commands comprise a relative root command $c_t^{\mathrm{js}}=[\Delta x_t,\Delta y_t,\Delta \mathrm{yaw}_t]$ and a binary drop command $b_t^{\mathrm{drop}}\in\{0,1\}$, derived during training from the reference trajectory and carry-end time (Appendix~\ref{sec:training-details}). In simulation, we use a nominal $37^\circ$ camera pitch, randomize camera pose, and inject depth noise, dropout, holes, edge artifacts, and offsets. The critic uses current robot, object, and reference states without history.

\noindent \textbf{Warm Start.}
We warm-start from a 23K-iteration checkpoint trained with the same method on box-only data, restoring only actor weights before training on all categories.
Training from scratch succeeds; warm starting accelerates convergence and is used for our released checkpoint.

\noindent \textbf{Training Details.}
We train 3-layer MLP teacher and student policies for 40K and 28K iterations, respectively, using 4096 environments per GPU on 8 NVIDIA L40S GPUs. Distillation uses teacher rollouts as motion references instead of the original kinematic trajectories. See Appendix~\ref{sec:training-details}.

\begin{table}[t]
\centering
\caption{\textbf{Cross-domain evaluation.} OMOMO-trained policies perform well on held-out OMOMO interactions but transfer poorly to PRISM.
PRISM-trained policies transfer to OMOMO and unseen PRISM-OOD categories; PRISM-ID evaluates the 80 training interactions.}
\resizebox{0.88\linewidth}{!}{
\begin{tabular}{l | c c c c}
\toprule
Training Data & OMOMO Train & OMOMO Test & PRISM ID & PRISM OOD \\
\midrule
OMOMO (90) & 94.44\% & 91.67\%  & 23.75\% &  12.50\%\\
PRISM ID(80) & \textbf{98.89\%} & \textbf{100\%} & \textbf{96.25\%} & \textbf{72.92\%} \\
\bottomrule
\end{tabular}
}
\vspace{-0.25cm}
\end{table}
\vspace{-0.25cm}
\section{Results}

\noindent \textbf{Overview.} 
We demonstrate that humanoid robots can learn  skills that generalize to diverse objects by imitating pure counterfactual generated videos. We first evaluate the effectiveness of our reconstructed data by comparing with the policy trained with OMOMO~\citep{li2023object} data only. Next, we evaluate the real-world performance over 20+ diverse objects. Finally, we conduct necessary ablation study to validate the necessity of each module in our framework design. All evaluations are conducted in MuJoCo~\citep{todorov2012mujoco} under a sim-to-sim setting, except for Sec.~\ref{sec:realworld} reports real-world robot experiments.
\vspace{-0.1cm}
\noindent \subsection{Reconstruction Quality from Real-to-Sim}

\noindent \textbf{Evaluation Data.}
We process and filter OMOMO to retain 63 high-quality human--object interaction sequences, then double the data by adding an object-scaled variant for each sequence using OmniRetarget~\citep{omniretarget}.
This gives 126 sequences, with 90 for \textit{OMOMO-Train} and 36 held out as \textit{OMOMO-Test}.
For PRISM, the reported comparison uses an earlier student distilled from 80 teacher rollouts and evaluates 80 \textit{PRISM-ID} interactions and 48 \textit{PRISM-OOD} interactions with chairs, tables, lamps, and monitors. Appendix~\ref{sec:data-details} distinguishes this evaluation from the updated training configuration.

\noindent \textbf{Evaluation Metrics.} For the distilled student policy, we derive joystick commands from the reference motion and render depth observations in simulation to drive the policy.
A rollout is successful if the robot completes pick--carry--drop and places the object at the desired target position.

\noindent \textbf{Discussion.}
OMOMO-trained policies perform well across OMOMO sequences, but transfer poorly to PRISM interactions, especially with out-of-domain objects. We identify three main failure modes: the policy struggles with unseen depth observations, overfits to close-range interactions and attempts to grasp before approaching distant objects, and fails to stabilize carry on novel object geometries. These failures suggest that OMOMO alone lacks the variation in perception, object configuration, and interaction dynamics required for robust loco-manipulation. In contrast, PRISM-trained policies generalize across both domains, demonstrating the broader transferability of PRISM data.

\vspace{-0.2cm}

\noindent \subsection{Real-World Deployment}
\label{sec:realworld}
We deploy our controller at 50\,Hz on a 29-DoF Unitree G1, with PD gains following~\citep{beyondmimic}. A head-mounted D435i camera captures stereo images at 30\,Hz, and Fast-FoundationStereo~\citep{wen2026fastfoundationstereo} estimates depth on an external computer connected to the robot via Ethernet. A human operator provides joystick commands. We freely place each object in front of the robot, varying its pose and distance across five trials. A trial succeeds if the robot reaches and grasps the object, then carries it stably for at least 3\,m without dropping it or falling. We evaluate all test objects (Fig.~\ref{fig:real_world_test_objects}) zero-shot, without using their scans or reconstructions for training. Videos are available on our \href{https://prism-real2sim2real.github.io/}{\textcolor{PRISMGreen}{project website}}.

\noindent \textbf{Discussion.}
We find that Fast-FoundationStereo substantially reduces the depth sim-to-real gap. We manually tilt the neck about $10^\circ$ upward from its default position, relying on camera-pose randomization during training to tolerate imprecise calibration. During bending, the neck sometimes resets to default or deviates from its target angle, producing out-of-distribution depth observations.

\vspace{-0.3cm}
\noindent  \subsection{Ablation Study}
\begin{figure}[t]
\centering
\includegraphics[width=\linewidth]{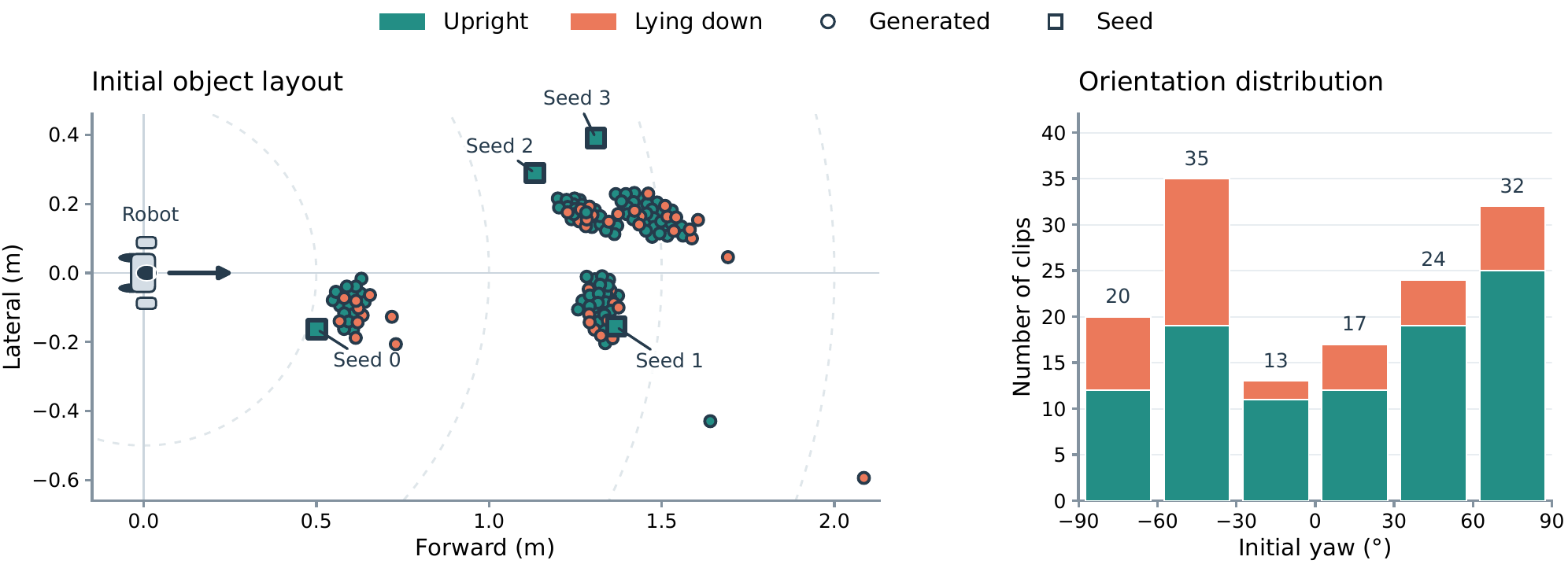}
\vspace{-0.35cm}
\caption{\textbf{Initial object-pose coverage.} Object poses relative to G1 in 137 generated clips and four upright seed videos. Left: planar positions of generated clips (circles) and seeds (squares). Right: pooled yaw counts in six $30^\circ$ bins modulo $180^\circ$. Colors distinguish upright and lying-down objects.}
\label{fig:object_pose_coverage}
\vspace{-0.35cm}
\end{figure}

Using eight seeds at submission, V2V outperforms geometric augmentation with fewer demonstrations (80 vs.\ 103; Appendix~\ref{sec:v2v-ablation}). We then ablate the key real-to-sim-to-real components on this expanded dataset. Figure~\ref{fig:object_pose_coverage} separately visualizes the initial object poses of 137 generated clips and four real-video seeds. Following~\citep{wang2025crisp}, a good reconstruction for robot learning should be physically plausible, simulatable and useful for policy training.  Therefore, we use the reconstructed trajectories to train policies in simulation, and report the downstream task success rate as the main metric. We define the baseline as: initialize object geometry and poses by SAM3D~\citep{sam3dteam2025sam3d3dfyimages}, then track it by FoundationPose~\citep{wen2024foundationposeunified6dpose}, and retarget it into simulation-ready robot--object demonstrations with OmniRetarget~\citep{omniretarget}. Table~\ref{tab:ablation} shows that better real-to-sim data directly benefits downstream policy learning. The baseline produces inaccurate robot--object interactions that limit effective policy learning. Anchored object pose and contact-aware retargeting progressively improve the data quality thus lead to stronger downstream performance. The contact reward further improves the policy by encouraging stable, task-relevant robot--object contacts during teacher and student training.
\vspace{-0.2cm}


\begin{table*}[t]
\centering
\caption{\textbf{Real-world evaluation.}
We evaluate our policy on real-world objects. Each in-domain category contains 3 different objects (see Fig.~\ref{fig:real_world_test_objects} for details). 
Each object is tested with 5 trials. }
\label{tab:real_world_eval_12cols}
\resizebox{\linewidth}{!}{
\begin{tabular}{l c c c c c c c c c c c c}
\toprule
& \multicolumn{4}{c}{\textbf{In-domain}} 
& \multicolumn{8}{c}{\textbf{Out-of-domain}} \\
\cmidrule(lr){2-5} \cmidrule(lr){6-13}
Metric 
& Ball 
& Bin 
& Barrel 
& Box 
& Paper towel 
& Helmet 
& Table 
& Backpack 
& Lamp 
& Chair 
& Kettle 
& Chick toy \\
\midrule
Success / Trials
& 12 / 15
& 14 / 15
& 14 / 15
& 15 / 15
& 5 / 5
& 3 / 5
& 5 / 5
& 5 / 5
& 3 / 5
& 4 / 5
& 4 / 5
& 3 / 5 \\
Success Rate
& 80\%
& 93\%
& 93\%
& 100\%
& 100\%
& 60\%
& 100\%
& 100\%
& 60\%
& 80\%
& 80\%
& 60\% \\
\bottomrule
\end{tabular}
}
\vspace{-0.5cm}
\end{table*}

\begin{table}[t]
\centering
\caption{\textbf{Ablation study.} We progressively add contact-anchored pose reconstruction, retargeting, and contact rewards. Each component improves task success on PRISM-ID and PRISM-OOD.}
\label{tab:ablation}
\resizebox{\linewidth}{!}{
\begin{tabular}{l c c c | c c}
\toprule
 & Contact-anchored Pose & Contact-anchored Retargeting & Contact Reward & Suc. (ID) & Suc. (OOD) \\
\midrule
Baseline
& \xmark & \xmark & \xmark
& 22.50\% & 12.50\% \\
+ anchored object pose
& \cmark & \xmark & \xmark
& 58.75\% & 29.17\% \\
+ contact-aware retargeting
& \cmark & \cmark & \xmark
& 86.25\% & 68.75\% \\
+ contact reward (PRISM)
& \cmark & \cmark & \cmark
& 96.25\% & 72.92\% \\
\bottomrule
\end{tabular}
}
\vspace{-0.47cm}
\end{table}

\begin{figure}[t]
\centering
\includegraphics[width=0.4975\linewidth,trim=3bp 188.974bp 2bp 2bp,clip]{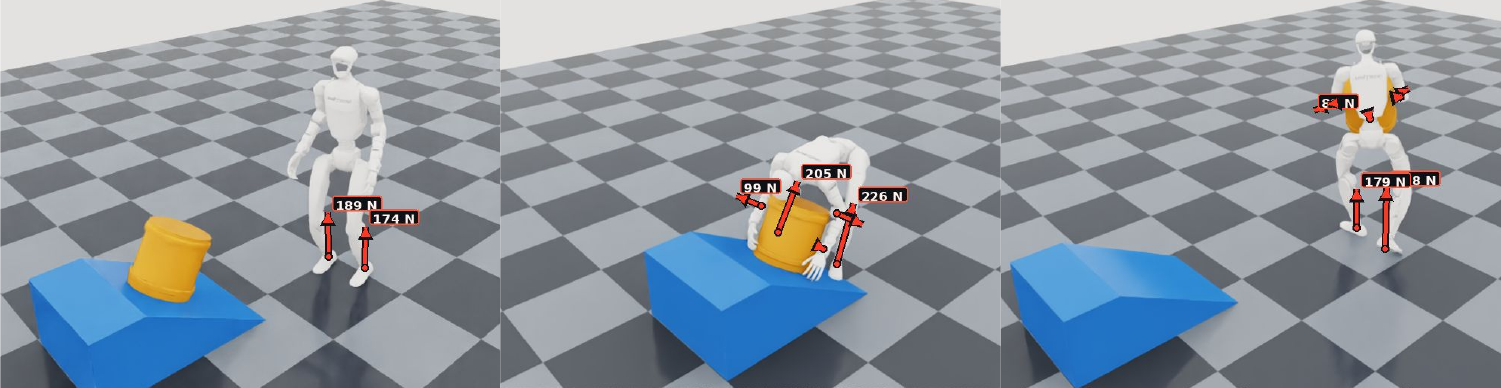}\hfill%
\includegraphics[width=0.4975\linewidth,trim=3bp 2.948bp 2bp 188.026bp,clip]{figs_1st/elevated_pickup.pdf}
\vspace{-0.45cm}
\caption{\textbf{Zero-shot elevated pick-up.} Though trained only on flat terrain, our policy picks up a box on a $35^\circ$ ramp ({\color{tile}\textit{left}}) and from a $0.43\,\mathrm{m}$ elevated support ({\color{tile}\textit{right}}), without additional policy tuning.}
\label{fig:elevated_pickup}
\vspace{-0.55cm}
\end{figure}

\vspace{0.2cm}
\noindent \textbf{Robustness and Generalization.} For reconstruction, we select seeds with stable views, limited occlusion, and task-compatible motions; V2V needs only simple prompts. At scale, heterogeneous-asset setup in Isaac Lab limits simulation and training throughput. Our policy succeeds zero-shot on a $35^\circ$ ramp and a $0.43\,\mathrm{m}$ elevated support (Fig.~\ref{fig:elevated_pickup}), likely due to grasp-height overlap with tall training objects, but fails at $45^\circ$ and $0.45\,\mathrm{m}$. Training on elevated V2V data may extend this range.

\vspace{-0.3cm}
\section{Conclusion}
\vspace{-0.2cm}
We presented \textsc{PRISM}, a real-to-sim-to-real framework that expands a few real videos into diverse humanoid demonstrations.
Contact-anchored reconstruction and retargeting turn imperfect counterfactual videos into physically plausible robot--object trajectories.
The resulting unified depth-based policy picks up, carries, and drops unseen objects zero-shot in the real world.
These results highlight video generation as a practical data source for generalizable whole-body interaction.

\vspace{-0.4cm}
\section{Limitations and Future Work}

Our pipeline delivers encouraging real-world results, yet several practical weaknesses remain.

\noindent \textbf{Reconstruction and Retargeting.}
Monocular 4D human--object--scene recovery remains challenging, especially when preserving consistent contacts and interaction dynamics.
Reconstruction errors can propagate through retargeting and compromise robot demonstrations.
Although Astra was unavailable during the development of this work, such coding agents could automate and iteratively refine reconstruction from counterfactual videos.
Combining these agents with contact-aware retargeting and simulation-based validation may improve the reliability of robot demonstrations.

\noindent \textbf{Data Scale.}
Four seed videos yield 256 generated clips, 137 feasible trajectories, and 129 successful teacher rollouts. This scale is insufficient to characterize scaling behavior, with reconstruction and retargeting remaining practical bottlenecks. Future work should jointly scale generation and trajectory recovery to study how demonstration quantity and quality affect zero-shot generalization.

\noindent \textbf{Object Physics.}
We use category-specific nominal masses and mesh-derived inertias, with coupled mass--inertia scaling and randomized friction (Appendix~\ref{sec:object-physics}). We do not identify instance-specific surface properties or internal mass distributions, which can cause sim-to-real contact mismatch.

\noindent \textbf{Simulation Limitations.}
We model dynamic objects as rigid bodies. Thus the policy can struggle with articulated or deformable objects whose contact geometry may change during interaction. For example, foldable chairs may shift through internal joints, causing unstable grasps or loss of control.

\acknowledgments{

We thank Chung Min Kim, Arthur Allshire, Hongsuk Choi, Isabella Yu, Junyi Zhang, Jacob Berg, Yen-Jen Wang, Sirui Chen, Charlie Cheng, Jiashun Wang, Siheng Zhao, Youjian Huang, JC Hu, Haochen Wang, Haozhi Qi and Qitao Zhao for their support and valuable feedback.
}


\bibliography{example}  

\newpage
\appendix
\section{Counterfactual Video Generation Details}
\label{sec: cf-gen}
We record four real-world seed videos of humans carrying boxes, selecting clips with clear full-body views, limited occlusion, and stable viewpoints. Using the SeedDance 2.0 web interface~\citep{seedance2026seedance}, we condition generation on each seed video and a category-level text prompt. The prompt replaces the manipulated object while preserving the background, lighting, camera viewpoint, and coarse task structure. These counterfactual videos depict interactions that could have occurred with different objects, allowing both object properties and human behavior to vary.

We generate 16 samples for each of four categories---boxes, bins, barrels, and balls---per seed video. This yields 64 counterfactual videos per seed and 256 videos in total. All generated videos are subsequently processed by our contact-anchored real-to-sim pipeline.

\noindent\textbf{Prompt template.} Figure~\ref{fig:v2v_prompt_template} shows the shared template. We instantiate \tok{CLS} with a target object category without specifying an individual instance's geometry, size, or pose, allowing the video model to generate variation within each category and adapt the human interaction accordingly.
\begin{figure}[ht]
\centering
\fbox{%
\begin{minipage}{0.86\linewidth}
\small\ttfamily\raggedright

\tok{VIDEO}

\vspace{0.6em}
In a real-world continuous footage. Preserve the reference video's original
background, lighting and camera viewpoint.

\vspace{0.6em}

{\setlength{\fboxsep}{3pt}
\colorbox{cfyellow}{%
\begin{minipage}{0.94\linewidth}
\small\ttfamily\raggedright
replace the box with \tok{CLS}, pick it up and carry with two hands.
\end{minipage}
}
}

\end{minipage}
}
\caption{\textbf{Video-to-video prompt template.}
\tok{VIDEO} denotes the seed video, and \tok{CLS} specifies a box, bin, barrel, or ball. The highlighted instruction changes the manipulated object while allowing the human interaction to adapt.}
\label{fig:v2v_prompt_template}
\end{figure}

\vspace{-0.35cm}
\section{Real-World Test Objects}
\label{sec:real-world-objects}
\begin{figure}[ht]
\centering
\begin{minipage}[t]{0.48\linewidth}
    \centering
    \textbf{In-Domain Objects}\\[0.3em]
    \includegraphics[width=\linewidth]{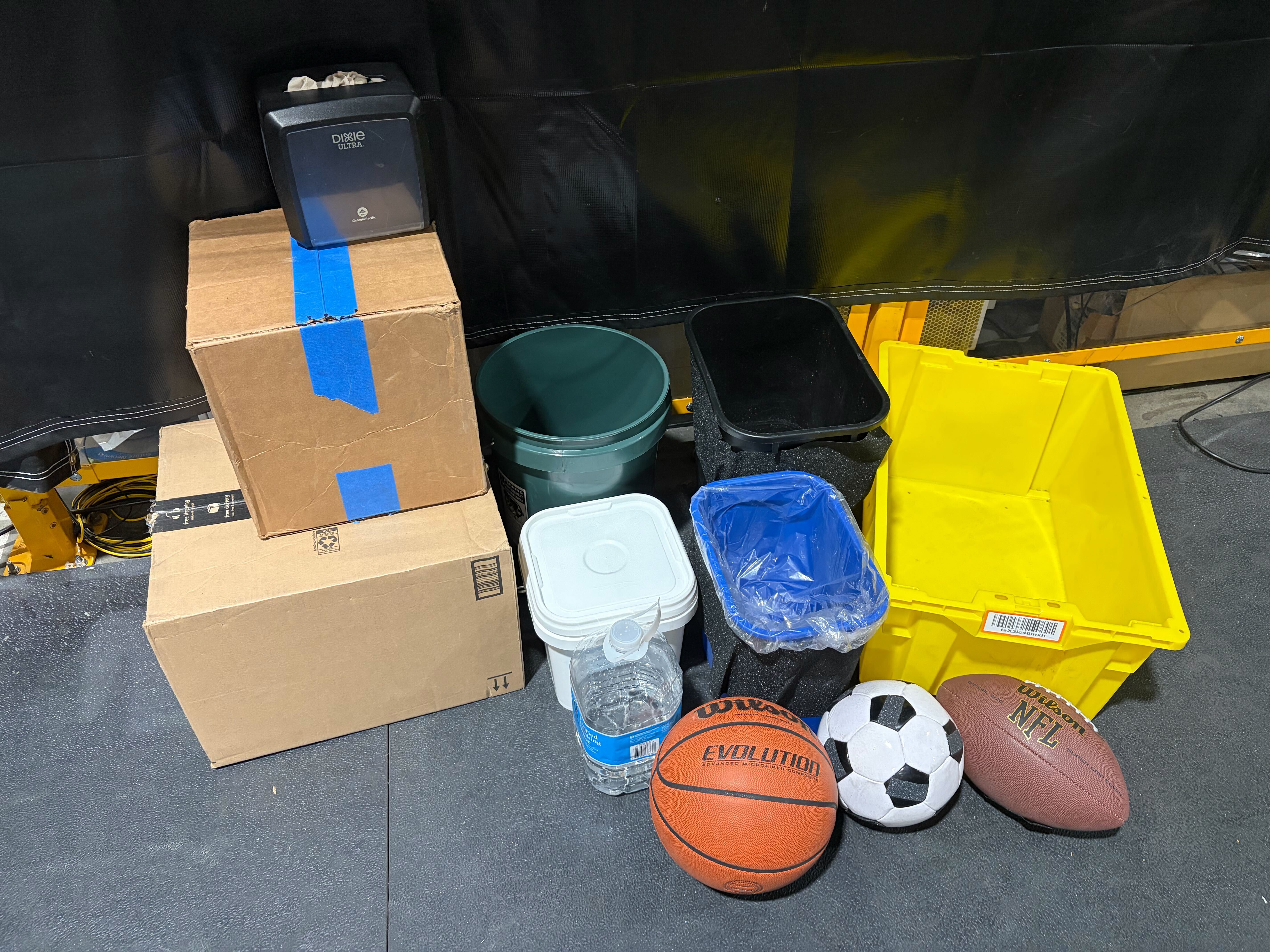}
\end{minipage}
\hfill
\begin{minipage}[t]{0.48\linewidth}
    \centering
    \textbf{Out-of-Domain Objects}\\[0.3em]
    \includegraphics[width=\linewidth]{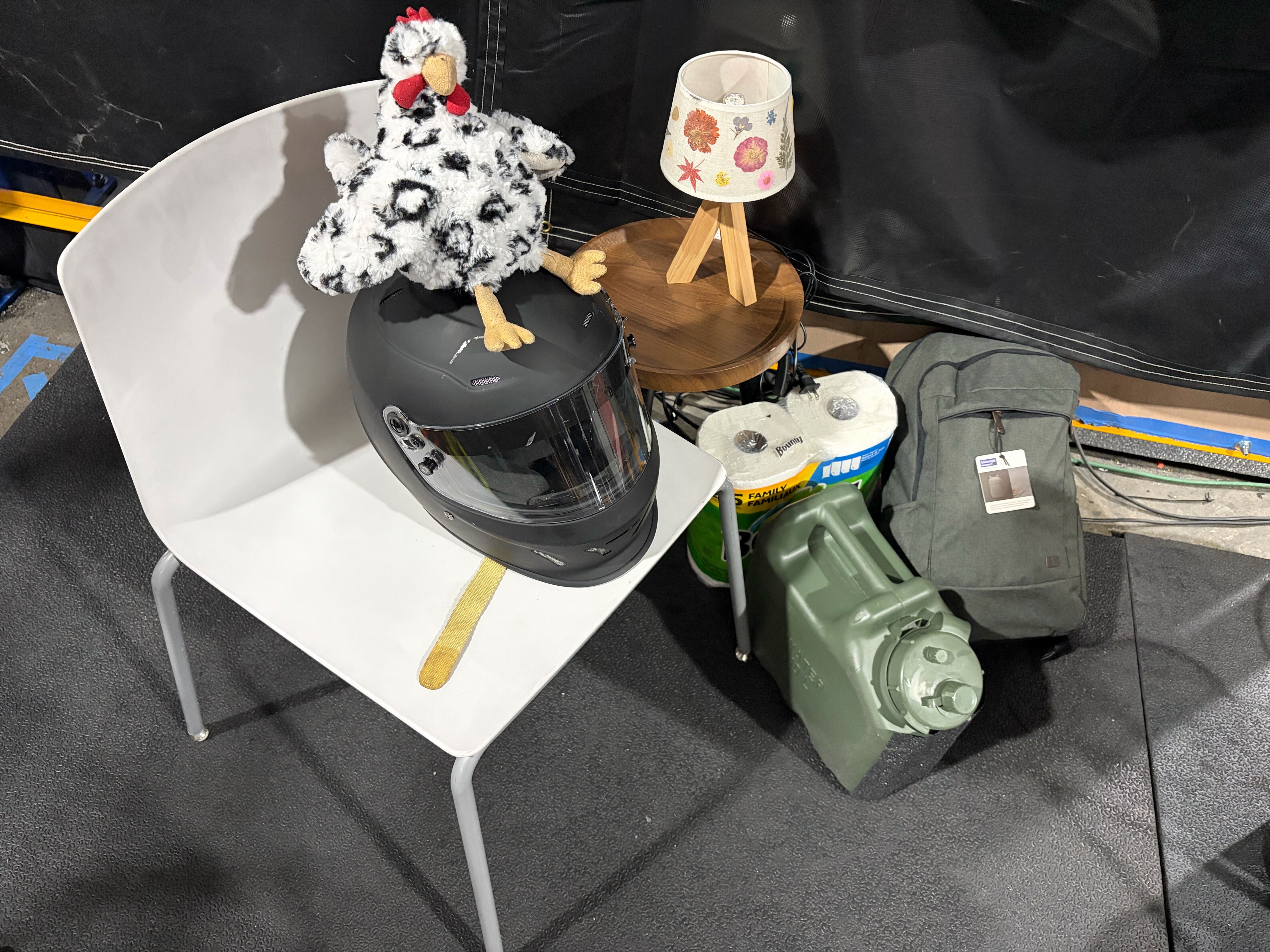}
\end{minipage}

\caption{\textbf{Real-world test objects.}
In-domain objects are unseen instances of boxes, bins, barrels, and balls. Out-of-domain objects belong to categories absent from the generated training videos. Both groups vary in appearance, geometry, weight, and scale.}
\label{fig:real_world_test_objects}
\end{figure}

We test unseen instances from the four generated object categories and objects from categories absent from the counterfactual videos (Fig.~\ref{fig:real_world_test_objects}). No scans or reconstructions of these real-world test objects are used for training. Following Sec.~\ref{sec:realworld}, each object is tested in five trials with varied initial poses and distances. A trial succeeds if the robot reaches and grasps the object, then carries it stably for at least 3\,m without dropping it or falling. Table~\ref{tab:real_world_eval_12cols} reports the resulting success rates.

\vspace*{0pt plus 1fill}

\section{Training Data and Simulation Evaluation}
\label{sec:data-details}
\noindent\textbf{Reconstruction and retargeting.} Of the 256 generated videos, 137 yield feasible robot--object trajectories through contact-anchored reconstruction and retargeting. The remaining sequences fail the constrained solver, mainly because of severe collisions. In particular, some generated objects are too large for the Unitree G1 humanoid to carry without object--body collisions.

We train the privileged co-tracking teacher on these 137 trajectories for 40K iterations and obtain 129 successful teacher rollouts. These rollouts provide the robot and object motion references for student distillation, replacing the original kinematic trajectories with interactions executed in simulation. We train the unified depth-based student for 28K iterations using this set of 129 demonstrations.

\noindent\textbf{Reported evaluation setting.} The cross-domain results in the main paper use an earlier student distilled from 80 successful rollouts of a 20K-iteration teacher checkpoint. Its PRISM-ID evaluation set comprises these 80 interactions. These results correspond to the earlier training configuration; results for the updated student trained on 129 demonstrations are not included in that comparison.

\noindent\textbf{Simulation evaluation data.} For the OMOMO comparison, we retain 63 human--object interaction sequences and add an object-scaled variant of each using OmniRetarget~\citep{omniretarget}. The resulting 126 sequences are split into 90 training sequences and 36 held-out test sequences. We additionally reconstruct 48 PRISM out-of-domain interactions, with 12 instances each from tables, chairs, lamps, and monitors. These held-out categories are absent from the generated training data.

\noindent\textbf{Simulation evaluation protocol.} We evaluate the distilled student in MuJoCo~\citep{todorov2012mujoco}, rendering depth observations and deriving joystick commands from the reference motion. Success requires completing pick--carry--drop and placing the object at the desired target position. This simulation metric includes the final placement, whereas the real-world metric above measures reaching, grasping, and stable carrying over at least 3\,m.

\vspace{-0.15cm}
\section{Training and Distillation Hyperparameters}
\vspace{-0.15cm}
\label{sec:training-details}
Both teacher and student policies use 3-layer MLPs and are trained for 40K and 28K iterations, respectively. Each stage runs 4096 environments per GPU on 8 NVIDIA L40S GPUs (32,768 environments in total). Tables~\ref{tab:actor_critic_observations}--\ref{tab:training_distillation_hyperparams} report observations, rewards, and hyperparameters. Object physics and payload sensitivity are detailed in Appendix~\ref{sec:object-physics}.

\noindent\textbf{Depth observations and deployment.} The student actor receives proprioception, joystick commands, and onboard depth. Joystick commands specify a relative planar root position and yaw, together with a binary drop command. During training, these are derived from the reference root trajectory and carry-end time. Depth is encoded by a small CNN into a 32-dimensional feature. At deployment, Fast-FoundationStereo~\citep{wen2026fastfoundationstereo} runs on an external computer linked to the robot via Ethernet, estimating depth from the head-mounted D435i stereo images. We find that this substantially reduces the sim-to-real gap in depth observations. The camera operates at 30\,Hz and the policy at 50\,Hz; the student transfers zero-shot without reference motion or external motion capture.

\noindent\textbf{Distillation and initialization.} We combine DAgger imitation with PPO, progressively increasing the PPO coefficient and retaining $\lambda=0.9$ during the final 20K iterations. Successful teacher rollouts supply the motion references. We first train a student for 23K iterations using the same method on box-only data. We then restore only its actor weights, leaving the critic and optimizer freshly initialized, and train for 28K iterations on boxes, bins, barrels, and balls. Training from scratch also succeeds; warm starting accelerates convergence and is used for our released checkpoint. We randomize camera pose and corrupt depth with noise, dropout, holes, edge artifacts, and offsets.

\noindent\textbf{Contact rewards.} Both teacher training and the student's PPO objective use object-frame contact anchors from reconstruction and retargeting. Rewards encourage reaching these targets with sufficient contact force. Table~\ref{tab:teacher_reward_weights} lists the teacher reward weights and scales.

\clearpage
\subsection{Observation and Reward Specifications}
\label{sec:observation-reward-tables}
\noindent\begin{minipage}{\linewidth}
  \centering
  \captionsetup{type=table}
\caption{\textbf{Observation spaces.}
  T/S denote teacher/student, and A/C denote actor/critic.
  \(\checkmark^{\dagger}\) denotes privileged student-critic inputs that are
  not used by the deployment policy.}
  \label{tab:actor_critic_observations}
  \begin{tabular}{@{}l c c c c c@{}}
    \toprule
    \textbf{Input} & \textbf{Dim.}
    & \textbf{T-A} & \textbf{T-C} & \textbf{S-A} & \textbf{S-C} \\
    \midrule

    \multicolumn{6}{@{}l}{\emph{Commands and references}} \\
    Motion command \(m_t\)                  & 58 & \(\checkmark\) & \(\checkmark\) & -- & \(\checkmark^{\dagger}\) \\
    Reference root position                 & 3  & -- & \(\checkmark\) & -- & \(\checkmark^{\dagger}\) \\
    Reference root orientation              & 6  & \(\checkmark\) & \(\checkmark\) & -- & \(\checkmark^{\dagger}\) \\
    Tracked body positions                  & 42 & -- & \(\checkmark\) & -- & \(\checkmark^{\dagger}\) \\
    Tracked body orientations               & 84 & -- & \(\checkmark\) & -- & \(\checkmark^{\dagger}\) \\
    Sparse root command \(c_t^{\mathrm{js}}\) & 3 & -- & -- & \(\checkmark\) & -- \\
    Drop button \(b_t^{\mathrm{drop}}\)     & 1  & -- & -- & \(\checkmark\) & -- \\

    \midrule
    \multicolumn{6}{@{}l}{\emph{Robot proprioception and actions}} \\
    Base linear velocity                    & 3  & -- & \(\checkmark\) & -- & \(\checkmark^{\dagger}\) \\
    Base angular velocity                   & 3  & \(\checkmark\) & \(\checkmark\) & \(\checkmark\) & \(\checkmark\) \\
    Joint positions                         & 29 & \(\checkmark\) & \(\checkmark\) & \(\checkmark\) & \(\checkmark\) \\
    Joint velocities                        & 29 & \(\checkmark\) & \(\checkmark\) & \(\checkmark\) & \(\checkmark\) \\
    Previous action                         & 29 & \(\checkmark\) & \(\checkmark\) & \(\checkmark\) & \(\checkmark\) \\

    \midrule
    \multicolumn{6}{@{}l}{\emph{Object state and perception}} \\
    Object pose \((p_o, R_o)\)              & 9  & \(\checkmark\) & \(\checkmark\) & -- & \(\checkmark^{\dagger}\) \\
    Target pose \((p_g, R_g)\)              & 9  & \(\checkmark\) & \(\checkmark\) & -- & \(\checkmark^{\dagger}\) \\
    Object size                             & 3  & \(\checkmark\) & \(\checkmark\) & -- & \(\checkmark^{\dagger}\) \\
    Object linear velocity                  & 3  & -- & \(\checkmark\) & -- & \(\checkmark^{\dagger}\) \\
    Object angular velocity                 & 3  & -- & -- & -- & \(\checkmark^{\dagger}\) \\
    Depth image / CNN latent                & \(58{\times}87 \rightarrow 32\)
                                               & -- & -- & \(\checkmark\) & -- \\

    \midrule
    \textbf{Total input} & --
      & \(175\)
      & \(310\)
      & \(94+32=126\)
      & \(313\) \\
    \textbf{Deployment} & --
      & train only
      & train only
      & \(\checkmark\)
      & train only \\

    \bottomrule
  \end{tabular}
\end{minipage}

\vspace{0.7cm}
\noindent\begin{minipage}{\linewidth}
\centering

\captionsetup{type=table}
\caption{\textbf{Main co-tracking reward weights for teacher training.}
Tracking terms use exponential kernels with the listed scale parameters.}
\label{tab:teacher_reward_weights}
\begin{tabular}{@{}p{0.38\textwidth}p{0.12\textwidth}p{0.42\textwidth}@{}}
\toprule
Reward term & Weight & Parameters \\
\midrule
Root position tracking & $0.5$ & $\sigma=0.3$ \\
Root orientation tracking & $0.5$ & $\sigma=0.4$ \\
Full-body position tracking & $1.0$ & $\sigma=0.3$ \\
Full-body orientation tracking & $1.0$ & $\sigma=0.4$ \\
Body linear velocity tracking & $1.0$ & $\sigma=1.0$ \\
Body angular velocity tracking & $1.0$ & $\sigma=3.14$ \\
Object position tracking & $1.0$ & $\sigma=0.3$ \\
Object orientation tracking & $1.0$ & $\sigma=0.4$ \\
Offline wrist target guidance & $5.0$ & Object-frame contact points; $\sigma=0.08$ \\
Offline force-gated contact guidance & $10.0$ & Force threshold $1.0$; force $\sigma=10.0$; contact schedule relaxation $5$ steps \\
Action-rate penalty & $-0.1$ & Squared action difference \\
Joint-limit penalty & $-10.0$ & Soft joint limit $0.9$ \\
Foot/ankle object-contact penalty & $-0.5$ & Threshold $1.0$ \\
Lower-body undesired contact penalty & $-0.1$ & Threshold $1.0$ \\
\bottomrule
\end{tabular}
\end{minipage}

\clearpage
\subsection{Optimization and Distillation Settings}
\vspace{-0.15cm}
\label{sec:optimization-settings}
\noindent\begin{minipage}{\linewidth}
\centering
\scriptsize
\setlength{\tabcolsep}{4pt}
\renewcommand{\arraystretch}{1.10}
\captionsetup{type=table}
\caption{\textbf{Training and distillation hyperparameters.}
Settings for the teacher and 129-rollout student; Appendix~\ref{sec:data-details} describes the earlier 80-rollout evaluation.}
\label{tab:training_distillation_hyperparams}
\vspace{-0.25cm}
\begin{tabular}{>{\raggedright\arraybackslash}p{0.23\textwidth}>{\raggedright\arraybackslash}p{0.34\textwidth}>{\raggedright\arraybackslash}p{0.34\textwidth}}
\hline
\textbf{Hyperparameter} & \textbf{Co-tracking teacher} & \textbf{Depth-based student} \\
\hline

Training data
& Contact-anchored retargeted trajectories
& Success-filtered teacher rollouts \\

Number of clips
& $137$
& $129$ \\

Total environments
& $4096 \times 8$
& $4096 \times 8$ \\

Policy input
& Motion command, robot state, object state
& Sparse root command, proprioception, buttons, depth \\

Object input
& Current pose, target pose, object size
& Depth observations; no explicit object state \\

History stacking
& None
& None \\

Actor network
& MLP $[512,256,128]$
& 3-layer MLP with depth latent input \\

Critic network
& MLP $[512,256,128]$
& MLP with privileged state \\

Activation
& ELU
& ELU \\

Depth input
& None
& $106 \times 60$ warped to $58 \times 87$ \\

Depth encoder
& None
& Small CNN, $32$-D latent \\

Camera range
& None
& $0.3$--$3.0$m \\

Camera pitch
& None
& $37^\circ$ \\

Depth noise
& None
& Hole prob. $0.2$, noise std. $0.03$m, offset std. $0.03$m \\

Rollout length / minibatches
& $24$ / $4$
& $24$ / $4$ \\

Learning epochs
& $7$
& $5$ \\

Optimizer
& AdamW
& AdamW \\

Actor / critic learning rate
& $1.0 \times 10^{-5}$ / $1.0 \times 10^{-5}$
& $7.0 \times 10^{-5}$ / $7.0 \times 10^{-5}$ \\

Discount factor
& $\gamma=0.99$
& $\gamma=0.99$ \\

GAE parameter
& $\lambda=0.95$
& $\lambda=0.95$ \\

PPO clip range
& $0.2$
& $0.2$ \\

Value loss coefficient
& $1.0$
& $1.0$ \\

Entropy coefficient
& $0.005$
& $0$ \\

Weight decay
& $10^{-3}$
& Not used \\

Max gradient norm
& $1.0$
& $1.0$ \\

Initial policy std.
& $1.0$
& $0.01$ \\

Training iterations
& $40{,}000$
& $28{,}000$ \\

Distillation loss
& None
& MSE behavior cloning \\

BC / DAgger coefficients
& None / None
& $1.0$ / $1.0$ \\

PPO coefficient schedule
& None
& $0.1 \rightarrow 0.9$, step $0.1$ every $500$ iterations \\

Teacher-action rollout mix
& None
& $0$ \\

Teacher action clip
& None
& $8.0$ \\

Contact-aware command
& None
& Peak-height mode, $\alpha=0.91$, smoothing $5$ steps \\

Sampling curriculum
& Success-rate adaptive clip weighting
& Adaptive timestep sampling  \\

Warm start
& --
& Box-only, 23K iterations; actor weights only \\

Reset schedule
& Random episode initialization
& Start-at-zero prob. $0.2 \rightarrow 1.0$ \\

Default-pose prepend
& $0.2$s
& Not used \\

Episode length
& Motion-dependent
& $8.0$s \\

Domain randomization
& Object physics (Appendix~\ref{sec:object-physics}), restitution, COM, joint bias, pushes
& Object physics (Appendix~\ref{sec:object-physics}), restitution, COM, joint bias, camera noise, pushes \\

External pushes
& Interval $[0.5,2.0]$s; max velocity $[0.7,0.7,0.25,0.7,0.7,1.0]$
& Interval $[0.5,2.0]$s; max velocity $[0.7,0.7,0.25,0.7,0.7,1.0]$ \\
\hline
\end{tabular}
\end{minipage}

\vspace{0.3cm}
\section{Object Physics and Payload Sensitivity}
\vspace{0.15cm}
\label{sec:object-physics}
Category-specific masses $m_0$ (Table~\ref{tab:object_physics_masses}) and mesh-derived inertias $\mathbf{I}_0$ scale as $(m,\mathbf{I})=s(m_0,\mathbf{I}_0)$, $s\sim\mathcal{U}(0.33,3.0)$. Shared friction is static $\mu_s\sim\mathcal{U}(0.1,0.7)$ and dynamic $\mu_d=r\mu_s$, with $r\sim\mathcal{U}(0.7,0.99)$. We do not identify instance-specific surface properties or mass distributions.

\noindent\begin{minipage}{\linewidth}
\vspace{0.25cm}
\centering
\small
\captionsetup{type=table}
\caption{\textbf{Object masses (kg).} Each mesh's inertia shares the mass scaling factor.}
\label{tab:object_physics_masses}
\begin{tabular*}{\linewidth}{@{}l@{\extracolsep{\fill}}cccc@{}}
\toprule
Category & Balls & Bins & Boxes & Barrels \\
\midrule
Nominal mass & $0.5$ & $1.0$ & $1.0$ & $1.5$ \\
Training range & $[0.165,1.50]$ & $[0.330,3.00]$ & $[0.330,3.00]$ & $[0.495,4.50]$ \\
\bottomrule
\end{tabular*}
\end{minipage}

\clearpage
\noindent\textbf{Payload sensitivity.}
Real-world test objects vary in shape and mass distribution, weighing $0.11\,\mathrm{kg}$ (box) to $4.99\,\mathrm{kg}$ (chair). Table~\ref{tab:payload_sensitivity} evaluates the earlier 80-rollout student (Appendix~\ref{sec:data-details}).

\noindent\begin{minipage}{\linewidth}
\vspace{-0.1cm}
\centering
\small
\captionsetup{type=table}
\caption{\textbf{Payload sensitivity.} Simulation success (\%) across five payload ranges.}
\label{tab:payload_sensitivity}
\vspace{-0.25cm}
\begin{tabular*}{\linewidth}{@{}l@{\extracolsep{\fill}}ccccc@{}}
\toprule
Payload (kg) & $0.1$--$1.0$ & $1.0$--$2.0$ & $2.0$--$3.0$ & $3.0$--$4.0$ & $4.0$--$5.0$ \\
\midrule
PRISM-ID  & $98.75$ & $97.50$ & $96.25$ & $93.75$ & $91.25$ \\
PRISM-OOD & $77.08$ & $72.92$ & $68.75$ & $62.50$ & $64.58$ \\
\bottomrule
\end{tabular*}
\end{minipage}
\vspace{-0.2cm}

\section{V2V versus Geometric Augmentation}
\label{sec:v2v-ablation}
These submission results use eight seed videos (Seed-1X). Seed-17X adds 16 yaw/$xy$/scale variants per seed ($8\times17=136$; 103 retained after depth-camera visibility filtering). All methods share training budgets and evaluation protocols. V2V uses the earlier 80-rollout student; these results do not evaluate the current four-seed, 129-rollout configuration.

\noindent\begin{minipage}{\linewidth}
\centering
\small
\captionsetup{type=table}
\caption{\textbf{V2V ablation at submission.} Success (\%); training sizes in parentheses.}
\label{tab:v2v_geometric_ablation}
\begin{tabular*}{\linewidth}{@{}l@{\extracolsep{\fill}}cccc@{}}
\toprule
Train $\backslash$ Eval & Seed-1X & Seed-17X & PRISM-ID & PRISM-OOD \\
\midrule
Seed-1X (8)   & $100.00$ & $58.25$ & $17.50$ & $6.25$ \\
Seed-17X (103) & $100.00$ & $98.06$ & $36.25$ & $22.92$ \\
V2V (80)      & $\mathbf{100.00}$ & $\mathbf{100.00}$ & $\mathbf{96.25}$ & $\mathbf{72.92}$ \\
\bottomrule
\end{tabular*}
\end{minipage}

V2V improves PRISM-ID/OOD performance with fewer demonstrations than Seed-17X.

\section{Additional Pipeline Analysis}
\label{sec:additional-analysis}
\subsection{Contact-Anchor Reliability}
\label{sec:contact-reliability}
For two-handed carrying, 2D overlap gives contact timing; mesh intersections with the segment between palm centers give object-frame anchors (Sec.~\ref{contact-a}). Pose/mesh errors can cause penetration and IK failure. We retain non-penetrating IK solutions and full-trajectory teacher rollouts (Appendix~\ref{sec:data-details}), without formal contact or feasibility guarantees.

\subsection{Pipeline Runtime}
\label{sec:pipeline-runtime}
Table~\ref{tab:pipeline_runtime} reports mean times per 8-second clip. Reconstruction and retargeting use one NVIDIA L40S GPU; V2V generation latency is recorded separately.

\noindent\begin{minipage}{\linewidth}
\centering
\small
\captionsetup{type=table}
\caption{\textbf{Pipeline runtime.} Mean processing time per 8-second clip.}
\label{tab:pipeline_runtime}
\begin{tabular*}{\linewidth}{@{}l@{\extracolsep{\fill}}ccc@{}}
\toprule
Stage & V2V generation & Reconstruction & Retargeting \\
\midrule
Mean time (s) & $191.76$ & $414.5$ & $71.6$ \\
\bottomrule
\end{tabular*}
\end{minipage}

\subsection{Additional Interaction Examples}
\label{sec:additional-interactions}
Figure~\ref{fig:additional_interactions} shows ball-rolling and door-opening reconstructions given suitable contact points and object representations. The door uses SAM3D~\citep{sam3dteam2025sam3d3dfyimages} for geometry and Codex for articulation. These qualitative examples are not policy or real-world evaluations; tool use, regrasping, and broader deformable-object manipulation and extreme dynamics as in ExoRecon~\citep{wang2025monofusion} remain untested.

\noindent\begin{minipage}{\linewidth}
\captionsetup{type=figure}
\centering
\includegraphics[width=\linewidth]{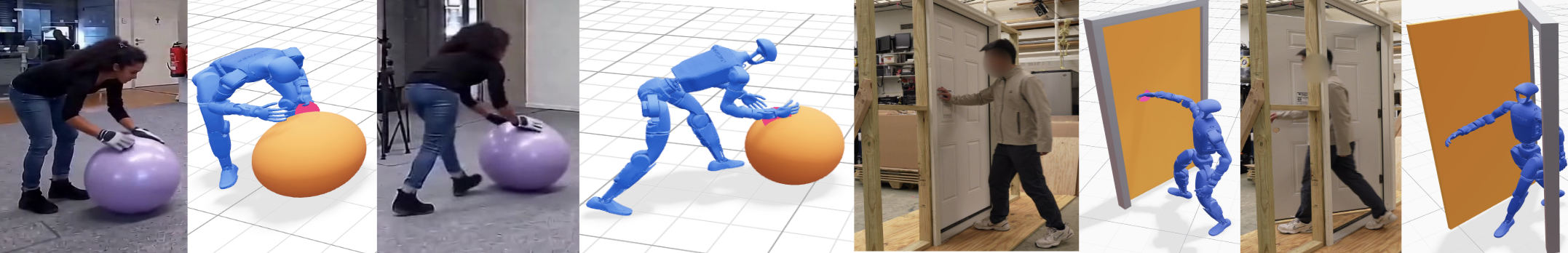}
\caption{\textbf{Additional interactions.} Input frames and reconstructions for ball rolling ({\color{tile}\textit{left}}) and door opening ({\color{tile}\textit{right}}), with human--object contacts highlighted.}
\label{fig:additional_interactions}
\end{minipage}

\end{document}